# Accuracy Measures for the Comparison of Classifiers

Vincent Labatut[1] and Hocine Cherifi[2]

[1] Galatasaray University, Computer Science Department,
Çırağan cad. n°36, 34357 İstanbul, Turkey
`vlabatut@gsu.edu.tr`

[2] University of Burgundy, LE2I UMR CNRS 5158,
Faculté des Sciences Mirande, 9 av. A. Savary, BP 47870, 21078 Dijon, France
`hocine.cherifi@u-bourgogne.fr`

*Abstract*— The selection of the best classification algorithm for a given dataset is a very widespread problem. It is also a complex one, in the sense it requires to make several important methodological choices. Among them, in this work we focus on the measure used to assess the classification performance and rank the algorithms. We present the most popular measures and discuss their properties. Despite the numerous measures proposed over the years, many of them turn out to be equivalent in this specific case. They can also lead to interpretation problems and be unsuitable for our purpose. Consequently, the classic overall success rate or marginal rates should be preferred for this specific task.

## I. INTRODUCTION

The comparison of classification algorithms is a complex and open problem. First, the notion of performance can be defined in many ways: accuracy, speed, cost, readability, etc. Second, an appropriate tool is necessary to quantify this performance. Third, a consistent method must be selected to compare the measured values.

Performance is most of the time expressed in terms of accuracy, which is why, in this work, we focus on this point. The number of accuracy measures appearing in the classification literature is extremely large. Some were specifically designed to compare classifiers [1], but most were initially defined for other purposes, such as measuring the association between two random variables [2], the agreement between two raters [3] or the similarity between two sets [4]. Furthermore, the same measure may have been independently developed by different authors, at different times, in different domains, for different purposes, leading to very confusing typology and terminology. Besides its purpose or name, what characterizes a measure is the definition of the concept of accuracy it relies on. Concretely, this means that measures are designed to focus on a specific aspect of the overall classification results [5]. This leads to measures with different interpretations, and some do not even have any clear interpretation. Finally, the measures may also differ in the nature of the situations they can be applied to: binary vs. multiclass problem, mutually exclusive classes vs. fuzzy classes [6], sampling design used to retrieve the data [7], etc.

Many different measures exist, but yet, there is no such thing as a perfect measure, which would be the best in every situation [8]: an appropriate measure must be chosen according to the classification context and objectives. Because of the overwhelming number of measures and of their heterogeneity, choosing the most adapted one is a difficult problem. Moreover, it is not always clear what the measures properties are, either because they were never rigorously studied, or because specialists do not agree on the question (e.g. the question of chance-correction [9]). Maybe for these reasons, authors very often select an accuracy measure by relying on the tradition or consensus observed in their field. The point is then more to use the same measure than their peers rather than the most appropriate one.

In this work, we reduce the complexity of choosing an accuracy measure by restraining our analysis to a very specific but widespread, situation. We discuss the case where one wants to select the best classification algorithm to process a given data set. An appropriate way to perform this task would be to study the data properties first, then to select a suitable classification algorithm and determine the most appropriate parameter values, and finally to use it to build the classifier. But not everyone has the statistical expertise required to perform this analytic work. Therefore, in practice, the most popular method consists in sampling a training set from the considered data, building various classifiers with different classification algorithms and parameters, and then comparing their performances empirically on some test sets sampled from the same data. Finally, the classifier with highest performance is selected and used on the rest of the data.

We will not address the question of the method used to compare performances. Instead, we will discuss the existing accuracy measures and their relevance to our specific context. We will be focusing on comparing classifiers with discrete outputs (by opposition to classifier outputting real scores or probabilities). Previous works already compared various measures, but with different purposes. Some authors adopted a general, context-independent approach and numerically compared measures by studying their correlation when applied to several real-world datasets [10, 11]. But as we underlined before, any comparison must be performed relatively to the application context. Other authors did focus on specific contexts, but these were different from our own. For instance, several studies of this kind exist in the remote sensing field [12, 13].

In the next section, we introduce the notations used in the rest of the paper. In section III, we review the main measures used as accuracy measures in the classification literature. Finally, in section IV, we compare and discuss their relevance relatively to our specific case.

## II. NOTATIONS AND TERMINOLOGY

Consider the problem of estimating $k$ classes for a test set containing $n$ instances. The true classes are noted $C_i$, whereas the *estimated* classes, as defined by the considered classifier, are noted $\hat{C}_i$ ($1 \leq i \leq k$).

Most measures are not processed directly from the raw classifier outputs, but from the *confusion matrix* built from these results. This matrix represents how the instances are distributed over estimated (rows) and true (columns) classes:

|   | $C_1$ | $\cdots$ | $C_k$ |
|---|---|---|---|
| $\hat{C}_1$ | $n_{11}$ | $\cdots$ | $n_{1k}$ |
| $\vdots$ | $\vdots$ | $\ddots$ | $\vdots$ |
| $\hat{C}_k$ | $n_{k1}$ | $\cdots$ | $n_{kk}$ |

The terms $n_{ij}$ ($1 \leq i,j \leq k$) correspond to the number of instances put in class number $i$ by the classifier (i.e. $\hat{C}_i$), when they actually belong to class number $j$ (i.e. $C_j$). Consequently, diagonal terms ($i = j$) correspond to correctly classified instances, whereas off-diagonal terms ($i \neq j$) represent incorrectly classified ones. Some measures are defined in terms of proportions rather than counts, so we additionally define the terms $p_{ij} = n_{ij}/n$.

The sums of the confusion matrix elements over row $i$ and column $j$ are noted $n_{i+}$ and $n_{+j}$, respectively. The corresponding sums of proportions, $p_{i+}$ and $p_{+j}$, are defined the same way.

When considering one class $i$ in particular, one may distinguish four types of instances: true positives (TP) and false positives (FP) are instances correctly and incorrectly classified as $\hat{C}_i$, whereas true negatives (TN) and false negatives (FN) are instances correctly and incorrectly not classified as $\hat{C}_i$, respectively. The corresponding counts are defined as $n_{TP} = n_{ii}$, $n_{FP} = n_{i,+} - n_{i,i}$, $n_{FN} = n_{+,i} - n_{i,i}$ and $n_{TN} = n - n_{TP} - n_{FP} - n_{FN}$, respectively. The corresponding proportions are noted $p_{TP}, p_{FP}, p_{FN}$ and $p_{TN}$, respectively.

## III. SELECTED MEASURES

### A. Nominal Association Measures

A measure of association is a numerical index, a single number, which describes the strength or magnitude of a relationship. Many association measures were used to assess classification accuracy, such as: chi-square-based measures ($\Phi$ coefficient, Pearson's $C$, Cramer's $V$, etc. [2]), Yule's coefficients, Matthew's correlation coefficient, Proportional reduction in error measures (Goodman & Kruskal's $\lambda$ and $\tau$, Theil's uncertainty coefficient, etc.), mutual information-based measures [14] and others. Association measures quantify how predictable a variable is when knowing the other one. They have been applied to classification accuracy assessment by considering these variables are defined by the distributions of instances over the true and estimated classes, respectively.

In our context, we consider the distribution of instances over estimated classes, and want to measure how much similar it is to their distribution over the true classes. The relationship assessed by an association measure is more general [2], since a high level of association only means it is possible to predict estimated classes when knowing the true ones (and vice-versa). In other terms, a high association does not necessary correspond to a match between estimated and true classes. For instance, if one considers a binary classification problem, both perfect classification and perfect misclassification give the same maximal association value.

Consequently, a confusion matrix can convey both a low accuracy and a high association at the same time, which makes association measures unsuitable for accuracy assessment.

### B. Overall Success Rate

Certainly the most popular measure for classification accuracy [15], the *overall success rate* is defined as the trace of the confusion matrix, divided by the total number $n$ of classified instances:

$$OSR = \frac{1}{n}\sum_{i=1}^{k} n_{i,i} \quad (1)$$

This measure is multiclass, symmetrical, and ranges from 0 (perfect misclassification) to 1 (perfect classification). Its popularity is certainly due to its simplicity, not only in terms of processing but also of interpretation, since it corresponds to the observed proportion of correctly classified instances.

### C. Marginal Rates

We gather under the term *marginal rates* a number of widely spread asymmetric class-specific measures. The *TP Rate* and *TN Rate* are both reference-oriented, i.e. they consider the confusion matrix columns (true classes). The former is also called sensitivity [15], producer's accuracy [13] and Dice's asymmetric index [16]. The latter is alternatively called specificity [15].

$$TPR_i = n_{TP}/(n_{TP} + n_{FN}) \quad (2)$$

$$TNR_i = n_{TN}/(n_{TN} + n_{FP}) \quad (3)$$

The correspondent estimation-oriented measures, which focus on the confusion matrix rows (estimated classes), are the *Positive Predictive Value* (PPV) and *Negative Predictive Value* (NPV) [15]. The former is also called precision [15], user's accuracy [13] and Dice's association index [16].

$$PPV_i = n_{TP}/(n_{TP} + n_{FP}) \quad (4)$$

$$NPV_i = n_{TN}/(n_{TN} + n_{FN}) \quad (5)$$

TNR and PPV are related to type I error (FP) whereas TPR and NPV are related to type II error (FN). All four measures range from 0 to 1, and their interpretation is straightforward. TPR (resp. TNR) corresponds to the proportion of instances belonging (resp. not belonging) to the considered class and actually classified as such. PPV (resp. NPV) corresponds to the proportion of instances predicted to belong (resp. not to belong) to the considered class, and which indeed do (resp. do not).

### D. F-measure and Jaccard Coefficient

The *F-measure* corresponds to the harmonic mean of PPV and TPR [15], therefore it is class-specific and symmetric. It is also known as Sørensen's similarity coefficient [17], Dice's coincidence index [16] and Hellden's mean accuracy index [18]:

$$F_i = 2 \frac{PPV_i \times TPR_i}{PPV_i + TPR_i} = \frac{2n_{TP}}{2n_{TP} + n_{FN} + n_{FP}} \quad (6)$$

It can be interpreted as a measure of overlapping between the true and estimated classes (other instances, i.e. TN, are ignored), ranging from 0 (no overlap at all to 1 (complete overlap).

The measure known as *Jaccard's coefficient of community* was initially defined to compare sets [4], too. It is a class-specific symmetric measure defined as:

$$JCC_i = \frac{n_{TP}}{n_{TP} + n_{FP} + n_{FN}} \quad (7)$$

It is alternatively called Short's measure [19]. For a given class, it can be interpreted as the ratio of the estimated and true classes intersection to their union (in terms of set cardinality). It ranges from 0 (no overlap) to 1 (complete overlap). It is linearly related to the F-measure [20]: $JCC_i = F_i/(2 - F_i)$, which is why we describe it in the same section.

### E. Classification Success Index

The *Individual Classification Success Index* (ICSI), is a class-specific symmetric measure defined for classification assessment purpose [1]:

$$\begin{aligned} ICSI_i &= 1 - (1 - PPV_i + 1 - TPR_i) \\ &= PPV_i + TPR_i - 1 \end{aligned} \quad (8)$$

The terms $1 - PPV_i$ and $1 - TPR_i$ correspond to the proportions of type I and II errors for the considered class, respectively. ICSI is hence one minus the sum of these errors. It ranges from $-1$ (both errors are maximal, i.e. 1) to 1 (both errors are minimal, i.e. 0), but the value 0 does not have any clear meaning. The measure is symmetric, and linearly related to the arithmetic mean of TPR and PPV, which is itself called *Kulczynski's measure* [21].

The *Classification Success Index* (CSI) is an overall measure defined by averaging ICSI over all classes [1].

### F. Agreement Coefficients

A family of chance-corrected inter-rater agreement coefficients has been widely used in the context of classification accuracy assessment. It relies on the following general formula:

$$A = \frac{P_o - P_e}{1 - P_e} \quad (9)$$

Where $P_o$ and $P_e$ are the observed and expected agreements, respectively. The idea is to consider the observed agreement as the result of an intended agreement and a chance agreement. In order to get the intended agreement, one must estimate the chance agreement and remove it from the observed one.

Most authors use $P_o = OSR$, but disagree on how the chance agreement should be formally defined, leading to different estimations of $P_e$. For his popular $\kappa$, Cohen used the product of the confusion matrix marginal proportions [3]: $P_e = \sum_i p_{i+} p_{+i}$. Scott's $\pi$ relies instead on the class proportions measured on the whole data set, noted $p_i$ [22]: $P_e = \sum_i (p_i)^2$. Various authors, including Maxwell [23] for his *Random Error* (MRE), made the assumption classes are evenly distributed: $P_e = 1/k$.

The problems of assessing inter-rater agreement and classifier accuracy are slightly different though. Indeed, in the former, the true class distribution is unknown, whereas in the latter it is completely known. Moreover, both raters are considered as equivalent and interchangeable, whereas a classifier is evaluated relatively to a reference (true classes), whose location in the confusion matrix matters. The presented corrections for chance are defined in function of these specific traits of the inter-rater agreement problem, which means they might not be relevant in our situation.

### G. Ground Truth Index

Türk's *Ground Truth Index* (GTI) is another chance-corrected measure, but this one was defined specially for classification accuracy assessment [24]. Türk supposes the classifier has two components: one is always correct, and the other is random, which means it is correct only sometimes. A proportion $\theta$ of the instances are supposed to be classified by the infallible classifier, the rest being processed randomly. The coefficient $\theta$ is iteratively estimated and its value is considered as a measure of accuracy, interpreted as the proportion of instances the classifier will always classify correctly, even when processing other data. Both class-specific and overall versions exist, depending on whether one considers $\theta$ to be the same for all classes or class-specific.

The way this measure handles chance correction is more adapted to classification than the agreement coefficients [20]. However, it has several limitations regarding the processed data: it cannot be used with less than three classes, or on perfectly classified data, and most of all it relies on the hypothesis of quasi-independence, which is rarely met in practice [11].

## IV. DISCUSSION

### A. Class Focus

As illustrated in the previous section, a measure can assess the accuracy for a specific class or over all classes. The former is adapted to situations where one is interested in a given class, or wants to conduct a class-by-class analysis of the classification results.

It is possible to define an overall measure by combining class-specific values measured for all classes, for example by averaging them, like in CSI. However, even if the considered class-specific measure has a clear meaning, it is difficult to give a straightforward interpretation to the resulting overall measure, other than in terms of combination of the class-specific values. Inversely, it is possible to use an overall measure to assess a given class accuracy, by merging all classes except the considered one [2]. In this case, the interpretation is straightforward though, and depends directly on the overall measure.

One generally uses a class-specific measure in order to distinguish classes in terms of importance. This is not possible with most basic overall measures, because they consider all classes to be equally important. Certain more sophisticated measures allow associating a weight to each class, though. However, a more flexible method makes this built-in feature redundant. It consists in associating a weight to each cell in the confusion matrix, and then using a regular (unweighted) overall measure [20]. This method allows distinguishing, in terms of importance, not only classes, but also any possible kind of classification error.

*B. Functional Relationships*

Some of the measures we presented are monotonically related, and this property takes a particular importance in our situation. Indeed, our goal is to sort classifiers depending on their performance on a given data set. If two measures are monotonically related, then the order will be the same for both measures. This makes the F-measure and Jaccard's coefficient similar to us, and so are the ICSI and Kulczynski's measure.

Moreover, it is well known that various combinations of two quantities can be sorted by increasing order, independently from the considered quantities: minimum, harmonic mean, geometric mean, arithmetic mean, quadratic mean, maximum [25]. If the quantities belong to $[0;1]$, we can even put their product at the beginning of the previous list, as the smallest combination. If we consider the presented measures, this means all combinations of the same marginal rates are the same to us. For instance, the sensitivity-precision product will always be smaller than the F-measure (harmonic mean), which in turn will always be smaller than Kulczynski's measure (arithmetic mean). It finally turns out all these measures are similar in our context: sensitivity-precision product, F-measure, Kulczynski's measure, Jaccard's coefficient and ICSI (and certainly others). Besides these combinations of TPR and PPV, this also holds for various measures corresponding to combinations of TPR and TNR, not presented here for space issues [15, 26].

*C. Range*

In the classification context, one can consider two extreme situations: perfect classification (i.e. diagonal confusion matrix) and perfect misclassification (i.e. all diagonal elements are zeros). The former should be associated to the upper bound of the accuracy measure, and the latter to its lower bound.

Measure bounds can either be fixed or depend on the processed data. The former is generally considered as a favorable trait [27], because it allows comparing values measured on different data sets without having to normalize them for scale matters. Moreover, having fixed bounds makes it easier to give an absolute interpretation of the measured features.

In our case, we want to compare classifiers evaluated on the same data. Furthermore, we are interested in their relative accuracies, i.e. we focus only on their relative differences. Consequently, this trait is not necessary. But it turns out most authors normalized their measures in order to give them fixed bounds (usually $[-1;1]$ or $[0;1]$). Note their exact values are of little importance, since any measure defined on a given interval can easily be rescaled to fit another one. Thus, several supposedly different measures are actually the same, but transposed to different scales [27].

*D. Interpretation*

Our goal is to compare classifiers on a given dataset, for which all we need is the measured accuracies. In other words, numerical values are enough to assess which classifier is the best on the considered data. But identifying the best classifier is useless if we do not know the criteria underlying this discrimination, i.e. if we are not able to interpret the measure. For instance, being the best in terms of PPV or TPR has a totally different meaning, since these measures focus on type I and II errors, respectively.

Among the measures used in the literature to assess classifiers accuracy, some have been designed analytically, in order to have a clear interpretation (e.g. Jaccard's coefficient [4]). Sometimes, this interpretation is questioned, or different alternatives exist, leading to several related measures (e.g. agreement coefficients). In some other cases, the measure is an *ad hoc* construct, which can be justified by practical constraints or observation, but may lack an actual interpretation (e.g. CSI). Finally, some measures are heterogeneous mixes of other measures, and have no direct meaning (e.g. the combination of OSR and marginal rates described in [28]). They can only be interpreted in terms of the measures forming them, and this is generally considered to be a difficult task.

*E. Correction for Chance*

Correcting measures for chance is still an open debate. First, authors disagree on the necessity of this correction, depending on the application context [7, 20]. In our case, we want to generalize the accuracy measured on a sample to the whole population. In other terms, we want to distinguish the proportion of success the algorithm will be able to reproduce on different data from the lucky guesses made on the testing sample, so this correction seems necessary.

Second, authors disagree on the nature of the correction term, as illustrated in our description of agreement coefficients. We can distinguish two kinds of corrections: those depending only on the true class distribution (e.g. Scott's and Maxwell's) and those depending also on the estimated class distribution (e.g. Cohen's and Türk's). The former is of little practical interest for us, because such a measure is linearly related to the OSR (the correction value being the same for every tested algorithm), and would therefore lead to the same ordering of algorithms. The second is more relevant, but there is still concern regarding how chance should be modeled. Indeed, lucky guesses depend completely on the algorithm behind the considered classifier. In other words, a very specific model would have to be designed for each algorithm in order to efficiently account for chance, which seems difficult or even impossible.

## V. CONCLUSION

In this work, we reviewed the main measures used for accuracy assessment, from a specific classification perspective. We consider the case where one wants to compare different classification algorithms by testing them on a given data sample, in order to determine which one will be the best on the sampled population.

In this situation, it turns out several traits of the measures are not relevant to discriminate them. First, all monotonically related measures are similar to us, because they all lead to the same ordering of algorithms. This notably discards a type of chance correction. Second, their range is of little importance, because we are considering relative values.

Moreover, a whole subset of measures associating weights to classes can be discarded, because a simpler method allows distinguishing classes in terms of importance while using an unweighted multiclass measure. Concerning chance-correction, it appears it is needed for our purpose; however no existing estimation for chance seems relevant. Finally, complex measures based on the combination of other measures are difficult or impossible to interpret correctly.

Under these conditions, we advise the user to choose the simplest measures, whose interpretation is straightforward. For overall accuracy assessment, the OSR seems to be the most adapted. If the focus has to be made on a specific class, we recommend using both the TPR and PPV, or a meaningful combination such as the F-measure. A weight matrix can be used to specify differences between classes or errors.

We plan to complete this work by focusing on the slightly different case of classifiers with real-valued output. This property allows using additional measures such as the area under the ROC curve and various error measures [15].

### ACKNOWLEDGMENT

The authors are grateful to the referees for their constructive comments, which helped improving the quality of this article.